\documentclass[10pt, conference, compsocconf]{IEEEtran}
\ifCLASSINFOpdf
  \usepackage[pdftex]{graphicx}
\else
\fi
\usepackage[cmex10]{amsmath}
\usepackage{amssymb}
\usepackage{booktabs}
\newcommand\norm[1]{\left\lVert#1\right\rVert}

\usepackage{algpseudocode}
\usepackage{algorithm}

\usepackage{graphicx}
\usepackage{caption}
\usepackage{subcaption}
\usepackage[square,sort,comma,numbers]{natbib}

\usepackage{url}

\begin{document}
%
\title{VIGAN: Missing View Imputation with Generative Adversarial Networks}


\author{\IEEEauthorblockN{Chao Shang, Aaron Palmer, Jiangwen Sun, Ko-Shin Chen, Jin Lu,  Jinbo Bi}
	\IEEEauthorblockA{Department of Computer Science and Engineering\\
		University of Connecticut\\
		Storrs, CT, USA\\
		\{chao.shang, aaron.palmer, jiangwen.sun, ko-shin.chen, jin.lu, jinbo.bi\}@uconn.edu}
}


%


\maketitle

\begin{abstract}
In an era when big data are becoming the norm, there is less concern with the quantity but more with the quality and completeness of the data. In many disciplines, data are collected from heterogeneous sources, resulting in multi-view or multi-modal datasets. The missing data problem has been challenging to address in multi-view data analysis. Especially, when certain samples miss an entire view of data, it creates the missing view problem. Classic multiple imputations or matrix completion methods are hardly effective here when no information can be based on in the specific view to impute data for such samples. The commonly-used simple method of removing samples with a missing view can dramatically reduce sample size, thus diminishing the statistical power of a subsequent analysis. In this paper, we propose a novel approach for view imputation via generative adversarial networks (GANs), which we name by VIGAN. This approach first treats each view as a separate domain and identifies domain-to-domain mappings via a GAN using randomly-sampled data from each view, and then employs a multi-modal denoising autoencoder (DAE) to reconstruct the missing view from the GAN outputs based on paired data across the views. Then, by optimizing the GAN and DAE jointly, our model enables the knowledge integration for domain mappings and view correspondences to effectively recover the missing view. Empirical results on benchmark datasets validate the VIGAN approach by comparing against the state of the art. The evaluation of VIGAN in a genetic study of substance use disorders further proves the effectiveness and usability of this approach in life science.

\end{abstract}


%
\IEEEpeerreviewmaketitle

\section{Introduction}

In many scientific domains, data can come from a multitude of diverse sources. A patient can be monitored simultaneously by multiple sensors in a home care system. In a genetic study, patients are assessed by their genotypes and their clinical symptoms. A web page can be represented by words on the page or by all the hyper-links pointing to it from other pages. Similarly, an image can be represented by the visual features extracted from it or by the text describing it. Each aspect of the data may offer a unique perspective to tackle the target problem. It brings up an important set of machine learning problems associated with the efficient utilization, modeling and integration of the heterogeneous data. In the era of big data, large quantities of such heterogeneous data have been accumulated in many domains. The proliferation of such data has facilitated knowledge discovery but also imposed great challenges on ensuring the quality or completeness of the data. The commonly-encountered missing data problem is what we cope with in this paper. 


There are distinct mechanisms to collect data from multiple aspects or sources. In multi-view data analysis, samples are characterized or viewed in multiple ways, thus creating multiple sets of input variables for the same sample. For instance, a genetic study of a complex disease may produce two data matrices respectively for genotypes and clinical symptoms, and the records in the two matrices are paired for each patient. In a dataset with three or more views, there exists a one-to-one mapping across the records of every view. In practice, it is however more common that data collected from different sources are for different samples, which leads to multi-modal data analysis. To study Alzheimer's disease, a US initiative collected neuroimages (a modality) for a sample of patients and brain signals such as electroencephalograms (another modality) for a different sample of patients, resulting in unpaired data. The integration of these datasets in a unified analysis requires different mathematical modeling from the multi-view data analysis because there is no longer a one-to-one mapping across the different modalities. This problem is also frequently referred to domain mapping or domain adaptation in various scenarios. The method that we propose herein can handle both the multi-view and multi-modal missing data problem.

\begin{figure}[htbp]
	\center
	\includegraphics[scale=.28]{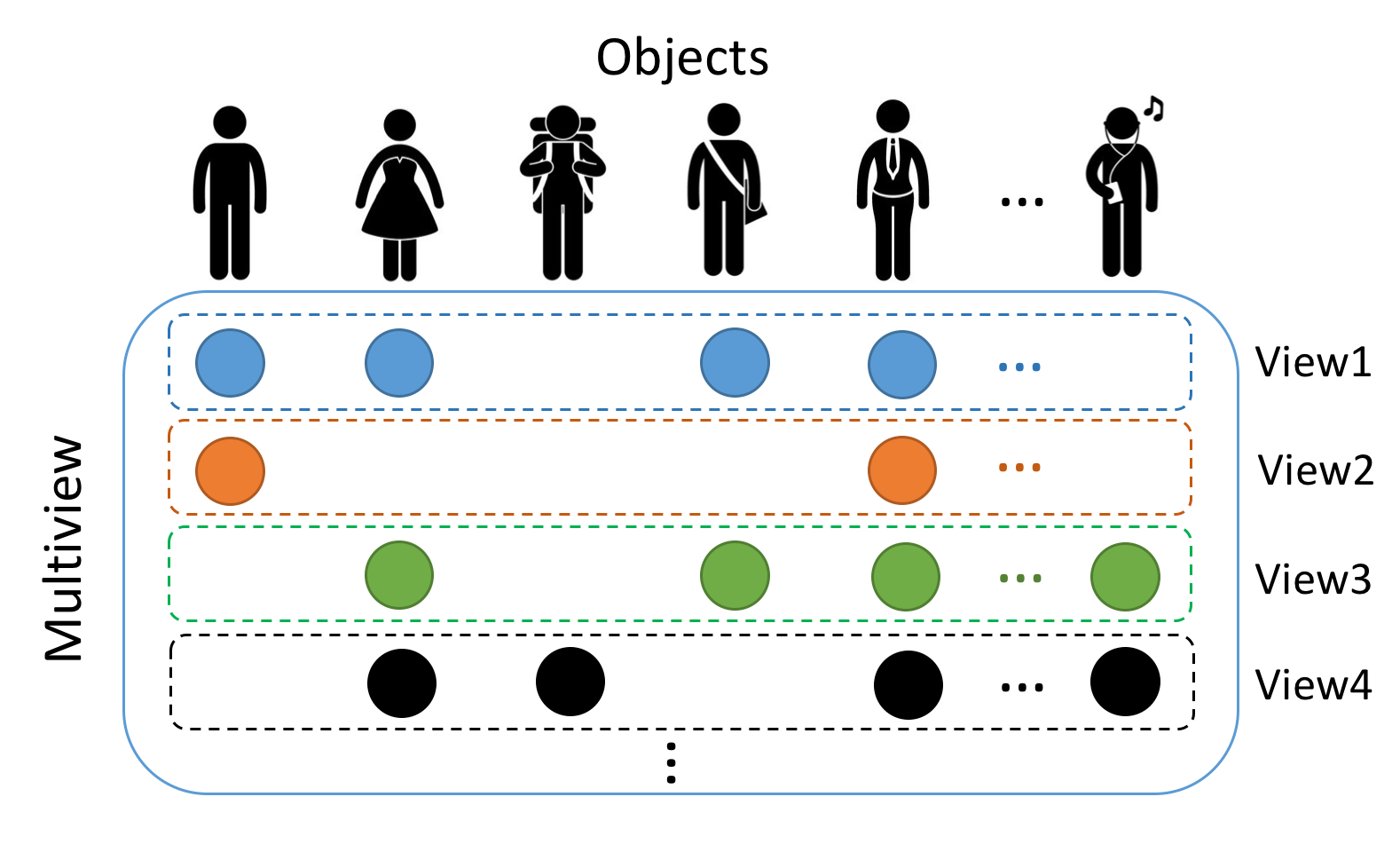}
	\caption{The missing view problem extremely limits the cross-view collaborative learning.} \label{fig:Fig1}
\end{figure}

Although the missing data problem is ubiquitous in large-scale datasets, most existing statistical or machine learning methods do not handle it and thus require the missing data to be imputed before the statistical methods can be applied \cite{troyanskaya2001missing,liew2010missing}. With the complex structure of heterogeneous data comes high complexity of missing data patterns. In the multi-view or multi-modal datasets, data can be missing at random in a single view (or modality) or in multiple views. Even though a few recent multi-view analytics \cite{trivedi2010multiview} can directly model incomplete data without imputation, they often assume that there exists at least one complete view, which is however often not the case. In multi-view data, certain subjects in a sample can miss an entire view of variables, resulting in the missing view problem as shown in Figure \ref{fig:Fig1}. In a general case, one could even consider that a multi-modal dataset just misses the entire view of data in a modality for the sample subjects that are characterized by another modality. 

To date, the widely-used data imputation methods focus on imputing or predicting the missing entries within a single view \cite{cai2010singular,candes2009exact,candes2010matrix}. Often times, data from multiple views are concatenated to form a single view data imputation problem. The classic single view imputation methods, such as multiple imputation methods, or matrix completion methods, are hardly scalable to big data. 
Lately, there has been research on imputation in true multi-view settings \cite{luo2015multiview,bhadra2017multi,williams2005analytical,hong2015multimodal, vincent2008extracting} where the missing values in a view can be imputed based on information from another complete view. These prior works assume that all views are available, and only some variables in each view are missing. This assumption has limited these methods because in practice it is common to miss an entire view of data for certain samples. This missing view problem brings up a significant challenge when conducting any multi-view analysis, especially when used in the context of very large and heterogeneous datasets like those in healthcare.

Recent deep learning methods \cite{wang2015deep, elkahky2015multi, ngiam2011multimodal} for learning a shared representation for multiple views of data have the potential to address the missing view problem. One of the most important advantages of these deep neural networks is their scalability and computational efficiency. Autoencoders \cite{hinton2006reducing} and denoising autoencoders (DAE) \cite{vincent2008extracting} have been used to denoise or complete data, especially for images. Generative adversarial networks (GANs) \cite{goodfellow2014generative} can create images or observations from random data sampled from a distribution, and hence can be potentially used to impute data. The latest GANs \cite{zhu2017unpaired, kim2017learning,yi2017dualgan, liu2016coupled,isola2016image} for domain mappings can learn the relationship between two modalities using unpaired data. However, all of these methods have not been thoroughly studied to impute missing views of data.

We propose a composite approach of GAN and autoencoder to address the missing view problem. Our method can impute an entire missing view by a multi-stage training procedure where in Stage one a multi-modal autoencoder \cite{ngiam2011multimodal} is trained on paired data to embed and reconstruct the input views.  Stage two consists of training a cycle-consistent GAN \cite{zhu2017unpaired} with unpaired data allowing a cross-domain relationship to be inferred.  Stage three re-optimizes both the pre-trained multi-modal autoencoder and the pre-trained cycle-consistent GAN so that we integrate the cross-domain relationship learned from unpaired data and the view correspondences learned from paired data.  Intuitively, the cycle-consistent GAN model learns to translate data between two views, and the translated data can be viewed as an initial estimate of the missing values, or a noisy version of the actual data. Then the last stage uses the autoencoder to refine the estimate by denoising the GAN outputs. 

There are several contributions in our approach:
\begin{enumerate}
	\item We propose an approach for the missing view problem in multi-view datasets.
	\item The proposed method can employ both paired multi-view data and unpaired multi-modal data simultaneously, and make use of all resources with missing data.
	\item Our approach is the first to combine domain mapping with cross-view imputation of missing data.
	\item Our approach is highly scalable, and can be extended to solve more than two views of missing data problem.
\end{enumerate}
Empirical evaluation of the proposed approach on both synthetic and real world datasets demonstrate its superior performance on data imputation and its computational efficiency. 
The rest of the paper will proceed as follows. In Section 2 we discuss related works. Section 3 is dedicated to the description of our method followed by a summary of experimental results in Section 4.  We then conclude in Section 5 with a discussion of future works.

\section{Related works}

\subsection{Matrix Completion}
Matrix completion methods focus on imputing the missing entries of a partially observed matrix under certain conditions. Specifically, the low-rank condition is the most widely used assumption, which is equivalent to assuming that each column of the matrix can be represented by a linear combination of a small number of basis vectors. Numerous matrix completion approaches have been proposed to complete a low-rank matrix, either based on convex optimization by minimizing the nuclear norm, such as the Singular Value Thresholding (SVT) \cite{cai2010singular} and SoftImpute \cite{mazumder2010spectral} methods, or alternatively in a non-convex optimization perspective by matrix factorization \cite{buchanan2005damped}. These methods are often ineffective when applied to the missing view problem. First, when concatenating features of different views in a multi-view dataset into a single data matrix, the missing entries are no longer randomly distributed, but rather appear in blocks, which violates the randomness assumption for most of the matrix completion methods. In this case, classical matrix completion methods no longer guarantee the recovery of missing data. Moreover, matrix completion methods are often computationally expensive and can become prohibitive for large datasets. For instance, those iteratively computing the singular value decomposition of an entire data matrix have a  complexity of $O(N^3)$ in terms of the matrix size $N$.

\subsection{Autoencoder and RBM}
Recently the autoencoder has shown to play a more fundamental role in the unsupervised learning setting for learning a latent data representation in deep architectures \cite{hinton2006reducing}.  Vincent et al introduced the denoising autoencoder in \cite{vincent2008extracting} as an extension of the classical autoencoder to use as a building block for deep networks.  

Researchers have extended the standard autoencoders into multi-modal autoencoders \cite{ngiam2011multimodal}.  Ngiam et al \cite{ngiam2011multimodal} use a deep autoencoder to learn relationships between high-level features of audio and video signals.  In their model they train a bi-modal deep autoencoder using modified but noisy audio and video datasets. Because many of their training samples only show in one of the modalities, the shared feature representations learned from paired examples in the hidden layers can capture correlations across different modalities, allowing for potential reconstruction of a missing view.  In practice, a multi-modal autoencoder is trained by simply zeroing out values in a view, estimating the removed values based on the counterpart in the other view, and comparing the network outputs and the removed values. Wang et al \cite{wang2015deep} enforce the feature representation of multi-view data to have high correlation between views.  
Another work \cite{srivastava2012multimodal} proposes to impute missing data in a modality by creating an autoencoder model out of stacked restricted Boltzmann machines. Unfortunately, all these methods train models from paired data.  During the training process, any data that have no complete views are removed, consequently leaving only a small percentage of data for training.

\subsection{Generative Adversarial Networks}
The method called generative adversarial networks (GANs) was proposed by Goodfellow et al \cite{goodfellow2014generative}, and achieved impressive results in a wide variety of problems. Briefly, the GAN model consists of a generator that takes a known distribution, usually some kind of normal or uniform distributions, and tries to map it to a data distribution.  The generated samples are then compared by a discriminator against real samples from the true data distribution.  The generator and discriminator play a minimax game where the generator tries to fool the discriminator, and the discriminator tries to distinguish between fake and true samples.  
Given the nature of GANs, they have great potential to be used for data imputation as further discussed in the next subsection of unsupervised domain mapping.  


\subsection{Unsupervised Domain Mapping}
Unsupervised domain mapping constructs and identifies a mapping between two modalities from unpaired data. 
There are several recent papers that perform similar tasks. DiscoGAN \cite{kim2017learning} created by Kim et al is able to discover cross-domain relations using an autoencoder model where the embedding corresponds to another domain.  A generator learns to map from one domain to another whereas a separate generator maps it back to the original domain.  Each domain has a discriminator to discern whether the generated images come from the true domain.  There is also a reconstruction loss to ensure a bijective mapping. Zhu et al use a cycle-consistent adversarial network, called CycleGAN \cite{zhu2017unpaired}, to train unpaired image-to-image translations in a very similar way.  Their architecture is defined slightly smaller because there is no coupling involved but rather a generated image is passed back over the original network. The pix2pix method \cite{isola2016image} is similar to the CycleGAN but trained only on paired data to learn a mapping from input to output images. Another method by Yi et al, callled DualGAN, uses uncoupled generators to perform image-to-image translation \cite{yi2017dualgan}. 

Liu and Tuzel coupled two GANs together in their CoGAN model \cite{liu2016coupled} for domain mapping with unpaired images in two domains. It is assumed that the two domains are similar in nature, which then motivates the use of the tied weights. 
Taigman et al introduce a domain transfer network in \cite{taigman2016unsupervised} which is able to learn a generative function that maps from one domain to another.  This model differs from the others in that the consistency they enforce is not only on the reconstruction but also on the embedding itself, and the resultant model is not bijective.  


\section{Method}
We now describe our imputation method for the missing view problem using generative adversarial networks which we call VIGAN. Our method combines two initialization steps to learn cross-domain relations from unpaired data in a CycleGAN and between-view correspondences from paired data in a DAE. Then our VIGAN method focuses on the joint optimization of both DAE and CycleGAN in the last stage.  
The denoising autoencoder is used to learn shared and private latent spaces for each view to better reconstruct the missing views, which amounts to denoise the GAN outputs. 

\begin{figure}[htbp]\
	\center
	\includegraphics[scale=.26]{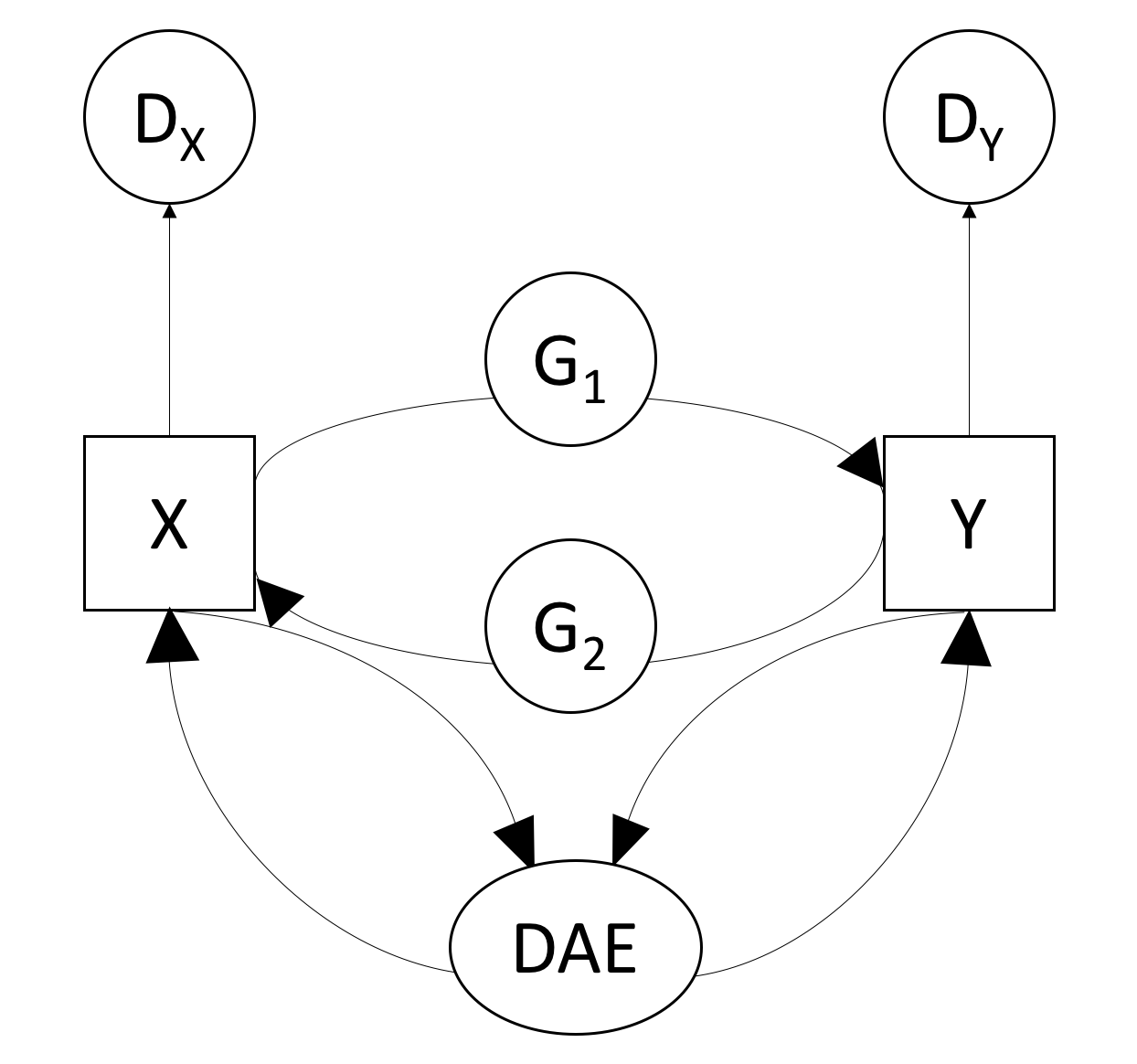}
	\caption{The VIGAN architecture consisting of the two main components: a CycleGAN with generators $G_1$ and $G_2$ and discriminators $D_X$ and $D_Y$ and a multi-modal denoising  autoencoder $DAE$.} \label{fig:Fig1_1}
\end{figure}


\subsection{Notations}
We assume that the dataset $\mathcal D$ consists of three parts: the complete pairs $\{(x^{(i)}, y^{(i)})\}_{i = 1}^N$, the $x$-only examples $\{x^{(i)}\}_{i = N+1}^{M_x}$, and the $y$-only examples $\{y^{(i)}\}_{i = N+1}^{M_y}$. We use the following notations.
\begin{itemize}
	\item $G_1: X \rightarrow Y \quad \mbox{and} \quad G_2: Y \rightarrow X$ are mappings between variable spaces $X$ and $Y$.
	\item $D_Y$ and $D_X$ are discriminators of $G_1$ and $G_2$ respectively. 
	\item $A: X \times Y \rightarrow X \times Y$ is an autoencoder function.
	\item We define two projections $P_X(x,y) = x$ and $P_Y(x,y) = y$ which either take the $x$ part or the $y$ part of the pair $(x,y)$.
	\item ${\mathbb E}_{x \sim p_{\text{data}}(x)}[f(x)] = \frac{1}{M_x} \sum_{i = 1}^{M_x} f(x^{(i)})$
	\item ${\mathbb E}_{(x,y) \sim p_{\text{data}}((x,y))}[f(x,y)] = \frac{1}{N} \sum_{i = 1}^{N} f(x^{(i)},y^{(i)})$
\end{itemize}

\subsection{The Proposed Formulation}
In this section we describe the VIGAN formulation which is also illustrated in Figure \ref{fig:Fig1_1}. Both paired and unpaired data are employed to learn mappings or correspondences between domains $X$ and $Y$. The denoising autoencoder is used to learn a shared representation from pairs $\{(x,y)\}$ and is pre-trained. The cycle-consistent GAN is used to learn from unpaired examples $\{x\}$, $\{y\}$ randomly drawn from the data to obtain maps between the domains. Although this mapping computes a $y$ value for an $x$ example (and vice versa), it is learned by focusing on domain translation, e.g. how to translate from audio to video, rather than finding the specific $y$ for that $x$ example. Hence, the GAN output can be treated as a rough estimate of the missing $y$ for an $x$ example. To jointly optimize both the DAE and CycleGAN, in the last stage, we minimize an overall loss function which we derive in the following subsections.

\noindent \textbf{The loss of multi-modal denoising autoencoder}

\begin{figure}[htbp]\
	\center
	\includegraphics[scale=.33]{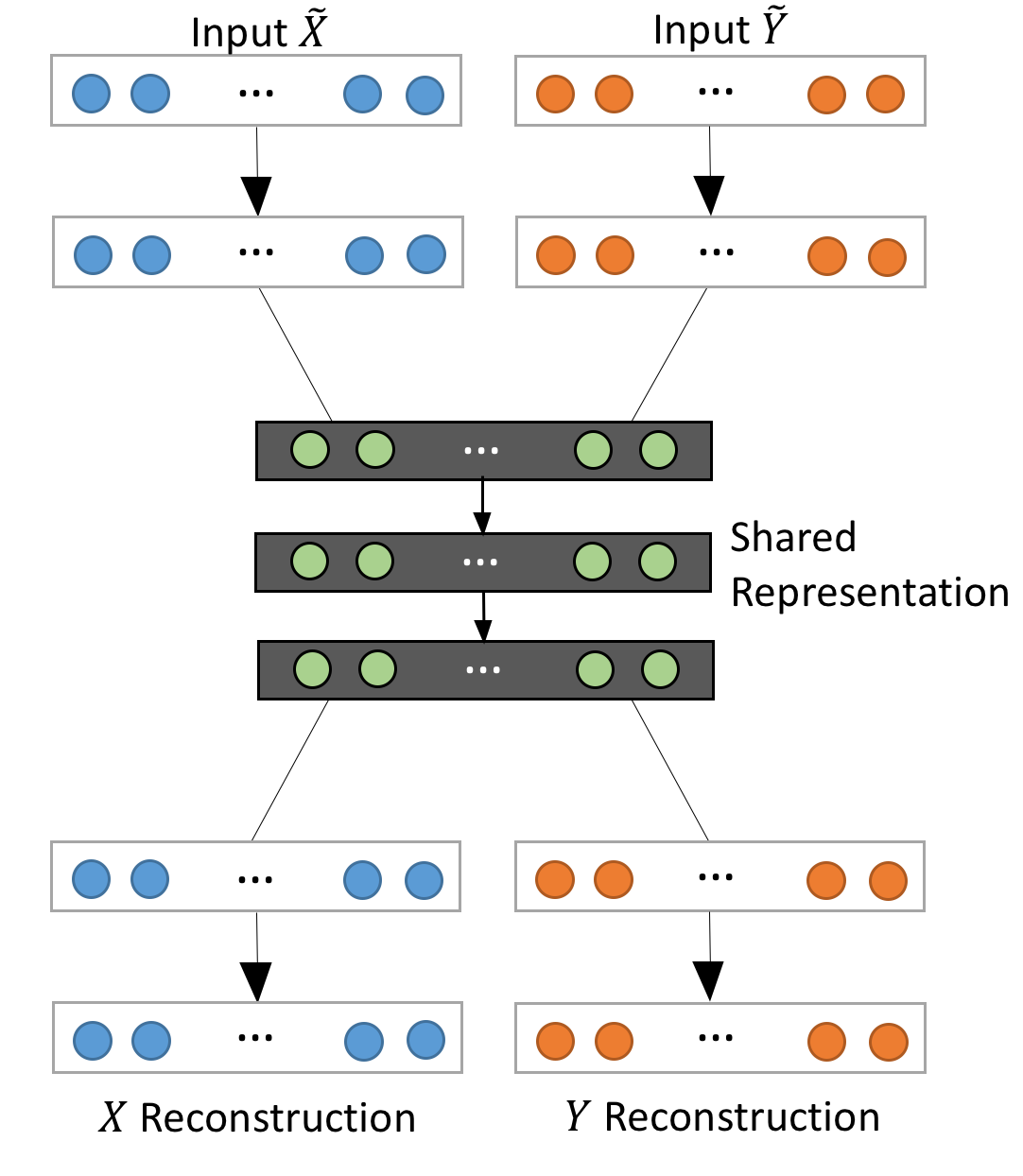}
	\caption{The multi-modal denoising autoencoder: the input pair $(\tilde{X},\tilde{Y})$ is $(x; G_1(x))$ or $(G_2(y); y)$ as corrupted (noising) versions of the original pair $(X; Y)$. } \label{fig:Fig3}
\end{figure}

The architecture of a multi-modal DAE consists of three pieces, as shown in Figure \ref{fig:Fig3}.  
The layers specific to a view will extract features from that view that will then be embedded in a shared representation as shown in the dark area in the middle of Figure \ref{fig:Fig3}. The shared representation is constructed by the layers that connect to both views. The last piece requires the network to reconstruct each of the views or modalities. The training mechanism aims to ensure that the inner representation catches the essential structure of the multi-view data. The reconstruction function for each view and the inner representation are jointly optimized. 

Given the mappings $G_1: X \rightarrow Y$ and $G_2: Y \rightarrow X$, we may view pairs $(x, G_1(x))$ and $(G_2(y), y)$ as two corrupted versions of the original pair $(x,y)$ in the data set. A denoising autoencoder, $A: X \times Y \rightarrow X \times Y$, is then trained to reconstruct $(x,y)$ from $(x, G_1(x))$ or $(G_2(y), y)$. We express the objective function as the squared loss:
\begin{align}
{\mathcal L}_{\text{AE}} & (A, G_1, G_2) = \notag\\
& {\mathbb E}_{(x,y) \sim p_{\text{data}}((x,y))}[ \norm{A(x, G_1(x)) - (x,y)}_2^2] \notag \\
& + {\mathbb E}_{(x,y) \sim p_{\text{data}}((x,y))}[ \norm{A(G_2(y), y) - (x,y)}_2^2].
\end{align}

\noindent \textbf{The adversarial loss}
\\
We then apply the adversarial loss introduced in \cite{goodfellow2014generative} to the composite functions $P_Y \circ A(x, G_1(x)): X \rightarrow Y$ and $P_X \circ A(G_2(y), y): Y \rightarrow X$.  This loss affects the training of both the autoencoder (AE) and the GAN so we name it $\mathcal L_{\text{AEGAN}}$, and it has two terms as follows: 
\begin{align}\label{eq:GANloss1}
& {\mathcal L}_{\text{AEGAN}}^{Y} (A , G_1, D_Y) = {\mathbb E}_{y \sim p_{\text{data}}(y)}[\log(D_Y (y))] \notag \\
& + {\mathbb E}_{x \sim p_{\text{data}}(x)}[\log(1 - D_Y (P_Y \circ A(x, G_1 (x))))],
\end{align}
and
\begin{align}\label{eq:GANloss2}
& {\mathcal L}_{\text{AEGAN}}^{X} (A , G_2, D_X) = {\mathbb E}_{x \sim p_{\text{data}}(x)}[\log(D_X (x))] \notag \\
& + {\mathbb E}_{y \sim p_{\text{data}}(y)}[\log(1 - D_X (P_X \circ A(G_2(y), y)))].
\end{align}

The first loss Eq.(\ref{eq:GANloss1}) aims to measure the difference between the observed $y$ value and the output of the composite function $P_Y \circ A(x, G_1 (x))$ whereas the second loss Eq.(\ref{eq:GANloss2}) measures the difference between the true $x$ value and the output of $P_X \circ A(G_2(y), y)$. The discriminators are designed to distinguish the fake data from the true observations. For instance, the $D_Y$ network is used to discriminate between the data created by $P_Y \circ A(x, G_1 (x))$ and the observed $y$. Hence, following the traditional GAN mechanism, we solve a minimax problem to optimize the parameters in $A$, $G_1$ and $D_Y$, i.e., $\min_{A, G_1} \max_{D_Y} {\mathcal L}_{\text{AEGAN}}^{Y}$. In alternating steps, we also solve $\min_{A, G_2} \max_{D_X} {\mathcal L}_{\text{AEGAN}}^{X}$ to optimize the parameters in the $A$, $G_2$ and $D_X$ networks.
Note that the above loss functions are used in the last stage of our method when optimizing both the DAE and GAN, which differs from the second stage of initializing the GAN where the standard GAN loss function ${\mathcal L}_{\text{GAN}}$ is used as discussed in CycleGAN \cite{zhu2017unpaired}.\\
\\
\textbf{The cycle consistency loss}
\\
Using a standard GAN, the network can map the same set of input images to any random permutation of images in the target domain. In other words, any mapping constructed by the network may induce an output distribution that matches the target distribution. Hence, the adversarial loss alone cannot guarantee that the constructed mapping can map an input to a desired output. To reduce the space of possible mapping functions, CycleGAN uses the so-called cycle consistency loss function expressed in terms of the $\ell_1$-norm penalty \cite{zhu2017unpaired}: 

\begin{align}
{\mathcal L}_{\text{CYC}} (G_1, G_2) = & {\mathbb E}_{x \sim p_{\text{data}}(x)}[ \norm{G_2 \circ G_1 (x) - x}_1] \notag \\
& + {\mathbb E}_{y \sim p_{\text{data}}(y)}[ \norm{G_1 \circ G_2 (y) - y}_1]
\end{align}

The rationale here is that by simultaneously minimizing the above loss and the GAN loss, the GAN network is able to map an input image back to itself by pushing through $G_1$ and $G_2$. This kind of cycle-consistent loss has been found to be important for a network to well perform as documented in CycleGAN \cite{zhu2017unpaired}, DualGAN \cite{yi2017dualgan}, and DiscoGAN \cite{kim2017learning}. By enforcing this additional loss, a GAN likely maps an $x$ example to its corresponding $y$ example in another view.
\\
\\
\textbf{The overall loss of VIGAN}
\\
After discussing the formulation used in the multi-modal DAE and CycleGAN, we are now ready to describe the overall objective function of VIGAN.
In the third stage of training, we formulate a loss function by taking into consideration all of the above losses as follows:
\begin{align} \label{eq:overall-loss}
{\mathcal L}&(A, G_1, G_2, D_X, D_Y) = \notag \\
& \lambda_{\text{AE}} {\mathcal L}_{\text{AE}} (A, G_1, G_2) +  \lambda_{\text{CYC}} {\mathcal L}_{\text{CYC}} (G_1, G_2) \notag \\
& + {\mathcal L}_{\text{AEGAN}}^{X} (A , G_2, D_X) +  {\mathcal L}_{\text{AEGAN}}^{Y} (A , G_1, D_Y) 
\end{align}
where $\lambda_{\text{AE}}$ and $\lambda_{\text{CYC}}$ are two hyper-parameters used to balance the different terms in the objective. We then solve the following minimax problem for the best parameter settings of the autoencoder $A$, generators $G_1$, $G_2$, and discriminators $D_X$ and $D_Y$:  
\begin{align}\label{eq:overall-problem}
\min_{A, G_1, G_2} & \max_{D_X, D_Y} {\mathcal L} (A, G_1, G_2, D_X, D_Y).
\end{align}

The overall loss in Eq.(\ref{eq:overall-loss}) uses both paired and unpaired data. In practice, even if all data are paired, the loss $\mathcal L_{\text{CYC}}$ is only concerned with the self-mapping. i.e., $x \rightarrow x$ or $y \rightarrow y$, and the loss $\mathcal L_{\text{AEGAN}}$ uses randomly-sampled $x$ or $y$ values, so both do not use the correspondence in pairs. Hence, Eq.(\ref{eq:overall-problem}) can still learn a GAN from unpaired data generated by random sampling from $x$ or $y$ examples. If all data are unpaired, the loss $\mathcal L_{\text{AE}}$ will degenerate to $0$, and the VIGAN can be regarded as an enhanced CycleGAN where the two generators $G_1$ and $G_2$ are expanded to both interact with a DAE which aims to denoise the $G_1$ and $G_2$ outputs for better estimation of the missing values (or more precisely the missing views). 

\subsection{Implementation}

\subsubsection{Training procedure}
As described above, we employ a multi-stage training regimen to train the complete model.  The VIGAN model first pre-trains the DAE where inputs are observed (true) paired samples from two views, which is different from the data used in the final step for the purpose of denoising the GAN. At this stage, the DAE is used as a regular multi-modal autoencoder to identify the correspondence between different views.
We train the multi-modal DAE for a pre-specified number of iterations. We then build the CycleGAN using unpaired data to learn domain mapping functions from view ${X}$ to view ${Y}$ and vice versa. 

\begin{figure}[htbp]\
	\center
	\includegraphics[scale=.35]{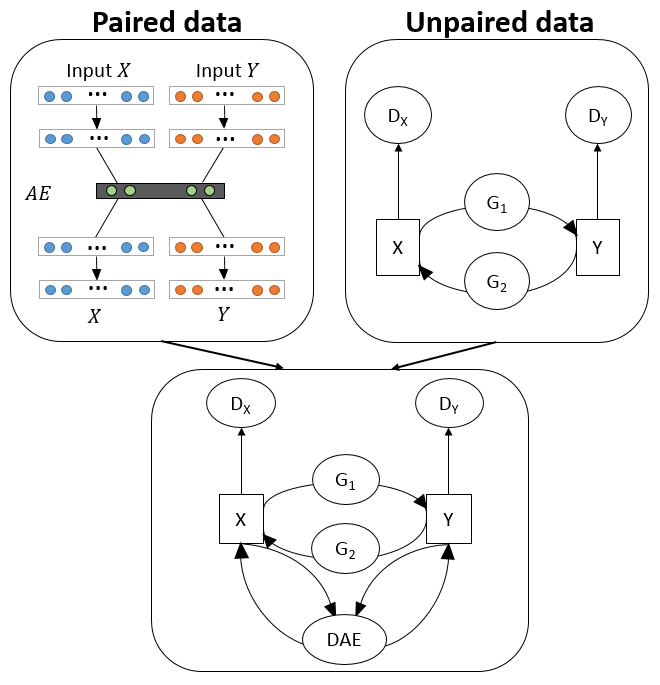}
	\caption{The multi-stage training process where the multi-modal autoencoder is first trained with paired data (top left).  The CycleGAN (top right) is trained with unpaired data.  Finally, these networks are combined into the final model and the training can continue with paired, unpaired or all data as needed.} \label{fig:Fig4}
\end{figure}

At last, the pre-trained DAE is re-optimized to denoise the outputs of GAN outputs by joint optimization with both paired and unpaired data. The DAE is now trained with the  noisy versions of $(x,y)$ as inputs, that are either $(x, G_1(x))$ or $(G_2(y), y)$, so the noise is added to only one component of the pair. The target output of the DAE is the true pair $(x,y)$. Because only one side of the pair is corrupted with certain noise (created by the GAN) in the DAE input, we aim to recover the correspondence by employing the observed counterpart in the pair. The difference from a regular DAE is that rather than corrupting the input with a noise of known distribution, we treat the residual of the GAN estimate as the noise.
This process is illustrated in Figure \ref{fig:Fig4} and the pseudo-code for the training is summarized in Algorithm \ref{alg: train}. There can be different training strategies. In our experiments, paired examples are used in the last step to refine the estimation of the missing views.

\begin{algorithm}[h]
	\caption{VIGAN training procedure}
	\label{alg: train}
	\begin{algorithmic}
		\State {\bfseries Require:} \\
		Image set $X$, image set $Y$, $n_1$ unpaired $x$ images $x_u^i$, $i=1,\cdots,n_1$ and $n_2$ unpaired y images $y_u^j$, $j=1,\cdots,n_2$, $m$ paired images $(x_p^k,y_p^k) \in X \times Y$, $k=1,\cdots,m$; The GAN generators for $x$ and $y$ have parameters $u_X$ and $u_Y$, respectively; the discriminators have parameters $v_X$ and $v_Y$; the DAE has parameters $w$; ${\mathcal L}(A)$ refers to the regular DAE loss; ${\mathcal L}(G_1, G_2, D_X, D_Y)$ refers to the regular CycleGAN loss; and ${\mathcal L}(A, G_1, G_2, D_X, D_Y)$ denotes the VIGAN loss.
	    \\
	    
		\textbf{Initialize} $w$ as follows:
		\textit{\\//Paired data}
			\For{the number of pre-specified iterations}  
			\State Sample paired images from $(x_p^k,y_p^k) \in X \times Y$
			\State Update $w$ to min  ${\mathcal L}(A)$
			\EndFor\\
		\textbf{Initialize} $v_X, v_Y, u_X, u_Y$ as follows:
		\textit{\\//Unpaired data}
		\For{the number of pre-specified iterations}  
		\State Sample unpaired images each from $x_u^i$ and $y_u^j$
		\State Update $v_X, v_Y$ to max ${\mathcal L}(G_1, G_2, D_X, D_Y)$
		\State Update $u_X, u_Y$ to min ${\mathcal L}(G_1, G_2, D_X, D_Y)$
		\EndFor
        \textit{\\//All samples or paired samples from all data}
		\For{the number of pre-specified iterations} 
		\State Sample paired images from $(x_p^k,y_p^k) \in X \times Y$ to form ${\mathcal L}_\text{AE}(A,G_1,G_2)$
		\State Sample from all images to form ${\mathcal L}_\text{AEGAN}$ and ${\mathcal L}_\text{CYC}$
		\State Update $v_X, v_Y$ to max ${\mathcal L}(A, G_1, G_2, D_X, D_Y)$
		\State Update $u_X, u_Y, w$ to min ${\mathcal L}(A,G_1,G_2,D_X,D_Y)$
		\EndFor
		
	\end{algorithmic}
\end{algorithm}

\subsubsection{Network architecture}
The network architecture may vary depending on whether we use numeric data or image data.  For example, we use regular fully connected layers when imputing numeric vectors, whereas we use convolutional layers when imputing images.  These are described in more detail in the following respective sections. 
\\
\\
\textbf{Network structure for numeric data:}
Our GANs for numeric data contain several fully connected layers.  A fully connected (FC) layer is one where a neuron in a layer is connected to every neuron in its preceding layer. Furthermore, these fully connected layers are sandwiched between the ReLU activation layers, which perform an element-wise ReLU transformation on the FC layer output.  The ReLU operation stands for rectified linear unit, and is defined as $max(0,z)$ for an input $z$.  The sigmoid layer is applied to the output layers of the generators, discriminators and the multi-modal DAE.

The multi-modal DAE architecture contains several fully connected layers which are sandwiched between the ReLU activation layers. Since we have two views in our multi-modal DAE, we concatenate these views together as an input to the network shown in Figure \ref{fig:Fig3}.  During training, the two views are connected in the hidden layers with the goal of minimizing the reconstruction error of both views.
\\
\\
\textbf{Network structure for image data:}
We adapt the architecture from the CycleGAN \cite{zhu2017unpaired} implementation which has shown impressive results for unpaired image-to-image translation.  The generator networks from \cite{zhu2017unpaired, johnson2016perceptual} contain two stride-2 convolutions, nine residual blocks \cite{he2016deep}, and two fractionally strided convolutions with stride 0.5. 
The discriminator networks use 70$\times$70 PatchGANs \cite{isola2016image, ledig2016photo, li2016precomputed}.  The sigmoid layer is applied to the output layers of the generators, discriminators and autoencoder to generate images within the desired range values.  The multi-modal DAE network \cite{ngiam2011multimodal} is similar to the numeric data architecture where the only difference is that we need to vectorize an image to form an input. Furthermore, the number of hidden nodes in these fully connected layers is changed from the original paper.

We used the adaptive moment (Adam) algorithm \cite{kingma2014adam} for training the model and set the learning rate to 0.0002.  All methods were implemented by PyTorch \cite{pytorch2017} and run on Ubuntu Linux 14.04 with NVIDIA Tesla K40C Graphics Processing Units (GPUs). Our code is publicly available at \url{https://github.com/chaoshangcs/VIGAN}.

\section{Experiments}

We evaluated the VIGAN method using three datasets, include MNIST, Cocaine-Opioid, Alcohol-Cannabis.  The Cocain-Opioid and Alcohol-Cannabis datasets came from an NIH-funded project which aimed to identify subtypes of dependence disorders on certain substances such as cocaine, opioid, or alcohol. To demonstrate the efficacy of our method and how to use the paired data and unpaired data for missing view imputation, we compared our method against a matrix completion method, a multi-modal autoencoder, the pix2pix and CycleGAN methods.  We trained the CycleGAN model using respectively paired data and unpaired data.

\subsection{Image benchmark data}

\textbf{MNIST dataset}
MNIST \cite{lecun1998mnist} is a widely known benchmark dataset consisting of 28 by 28 pixel black and white images of handwritten digits.  The MNIST database consists of a training set of 60,000 examples and a test set of 10,000 examples. We created a validation set by splitting the original training set into a new training set consisting of 54,000 examples and a validation set of 6,000 examples.  

Since this dataset did not have multiple views, we created a separate view following the method in the CoGAN paper where the authors created a new digit image from an original MNIST image by only maintaining the edge of the number \cite{liu2016coupled}. We used the original digit as the first view, whereas the second view consisted of the edge images.  We trained the VIGAN network assuming either view can be completely missing.
In addition, we divided the 60,000 examples into two equal sized disjoint sets as the unpaired datasets. The original images remained in one dataset, and the edge images were in another set.

\begin{figure}[!tbp]
	\begin{subfigure}[b]{0.235\textwidth}
		\includegraphics[width=\linewidth]{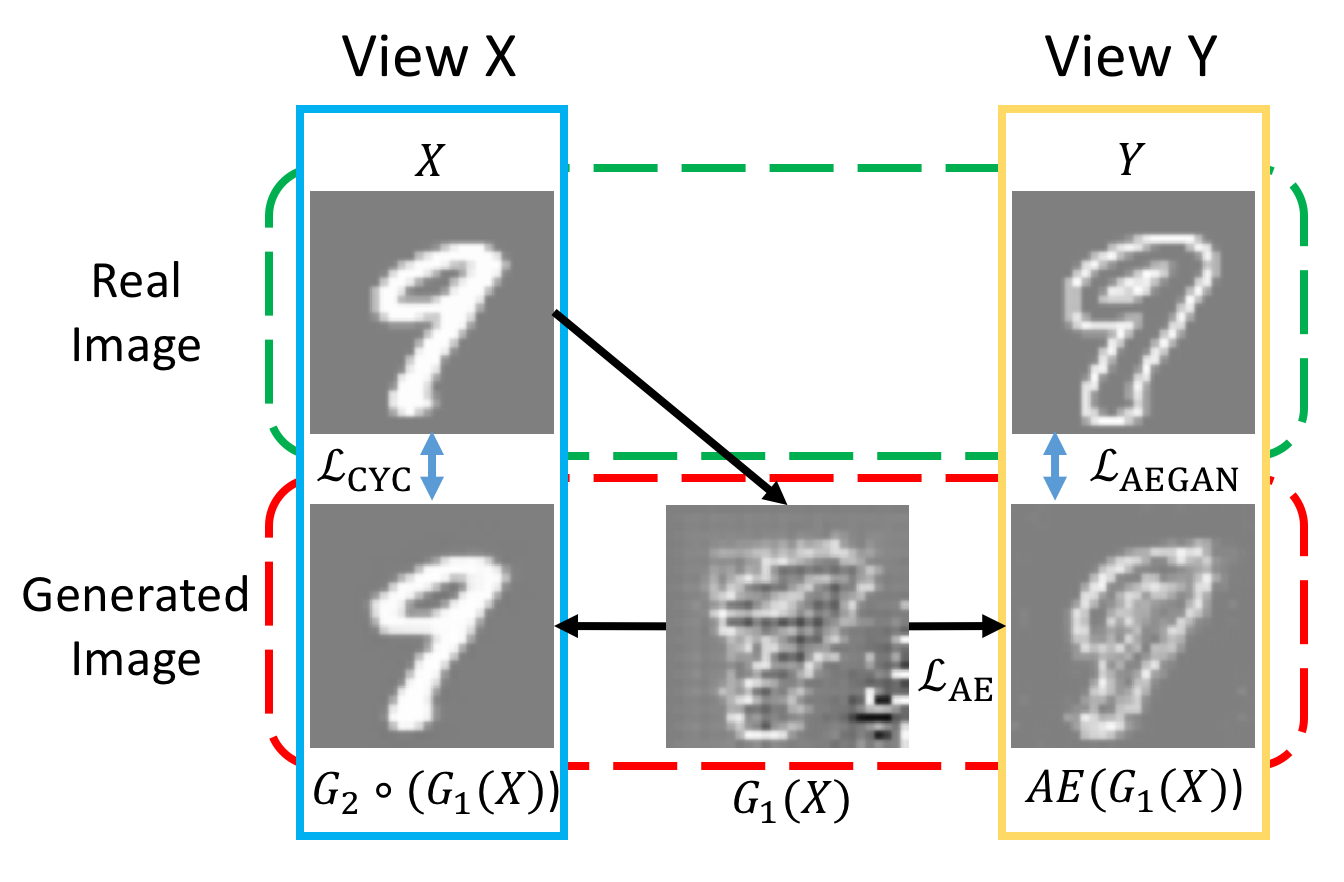}
		\caption{${X} \to {Y}$} 
	\end{subfigure}
	\begin{subfigure}[b]{0.235\textwidth}
		\includegraphics[width=\linewidth]{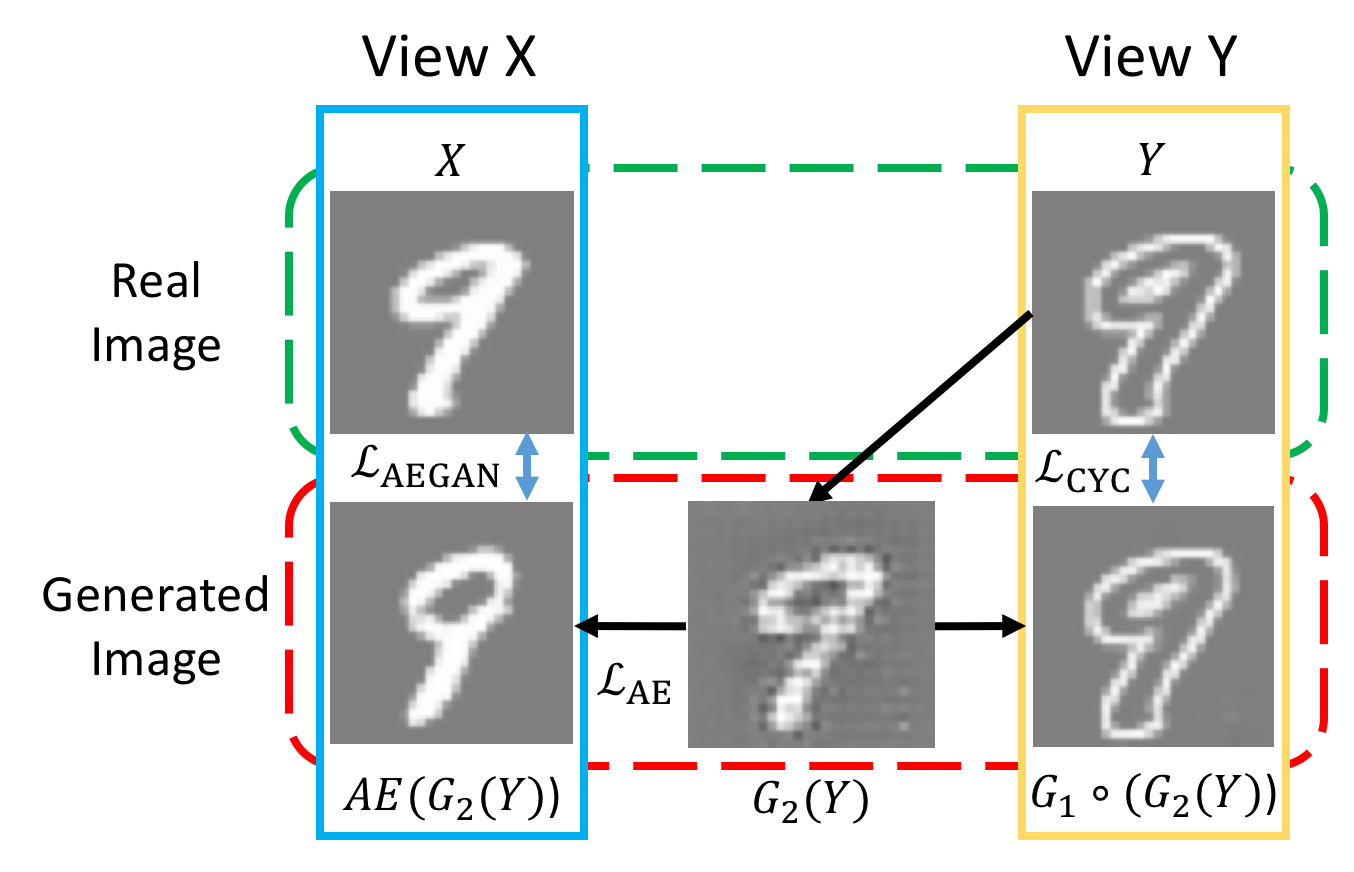}
		\caption{${Y} \to {X}$}
	\end{subfigure}
	
		\caption{The imputation examples.} \label{fig:Fige}
\end{figure}

Figure \ref{fig:Fige} demonstrates the results. It shows the imputed $y$ image in (a) where $G_1(x)$ is the initial estimate via the domain mapping.  The image labeled by $AE(G_1(X))$ is the denoised estimate, which gives the final imputed output. Figure \ref{fig:Fige}(b) shows the other way around.

\begin{figure}
	\begin{subfigure}[b]{0.25\textwidth}
		\includegraphics[width=\linewidth]{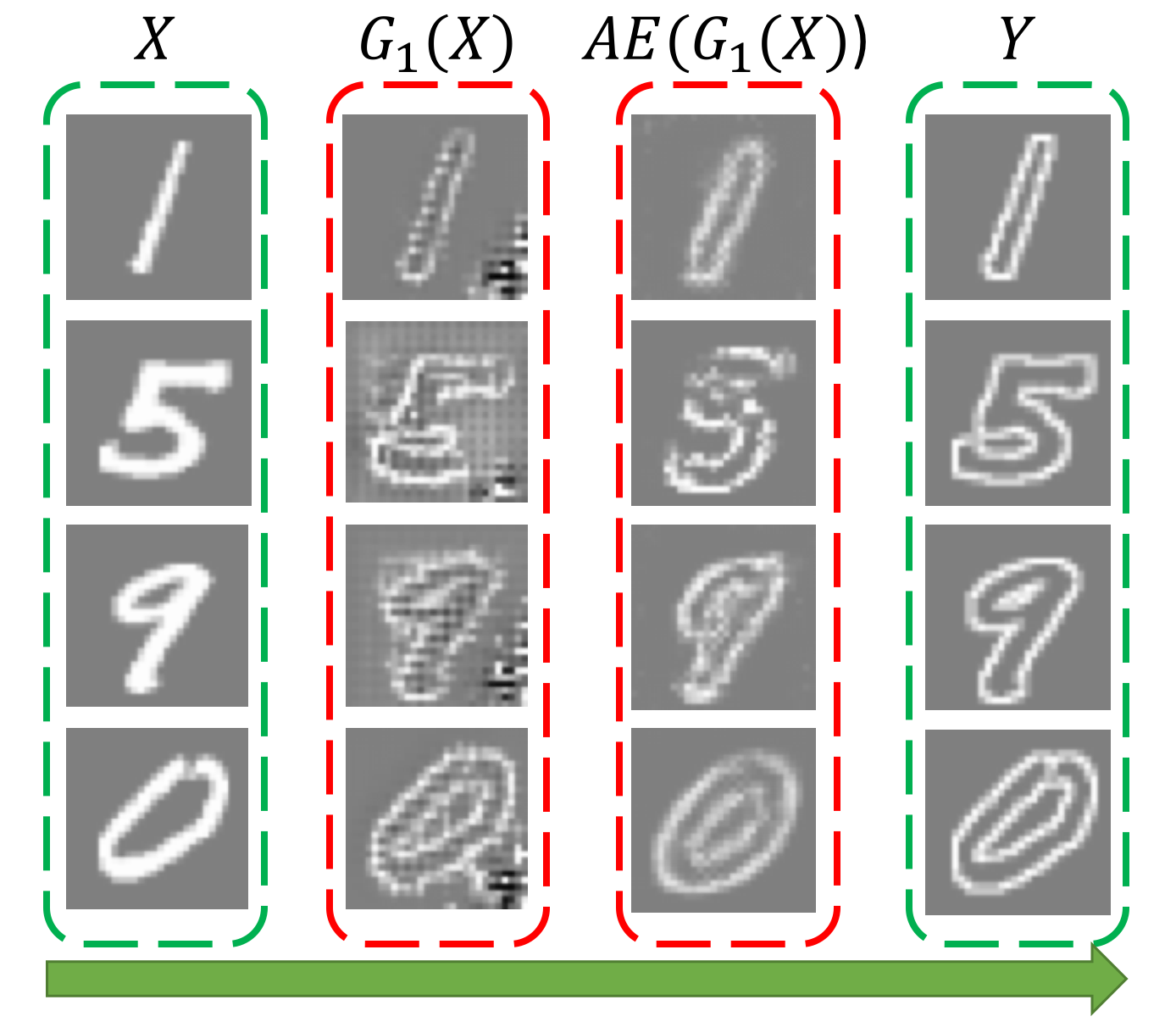}
		\caption{Outputs from ${X}$ to ${Y}$.}
		\label{fig:gull}
	\end{subfigure}%
	\begin{subfigure}[b]{0.25\textwidth}
		\includegraphics[width=\linewidth]{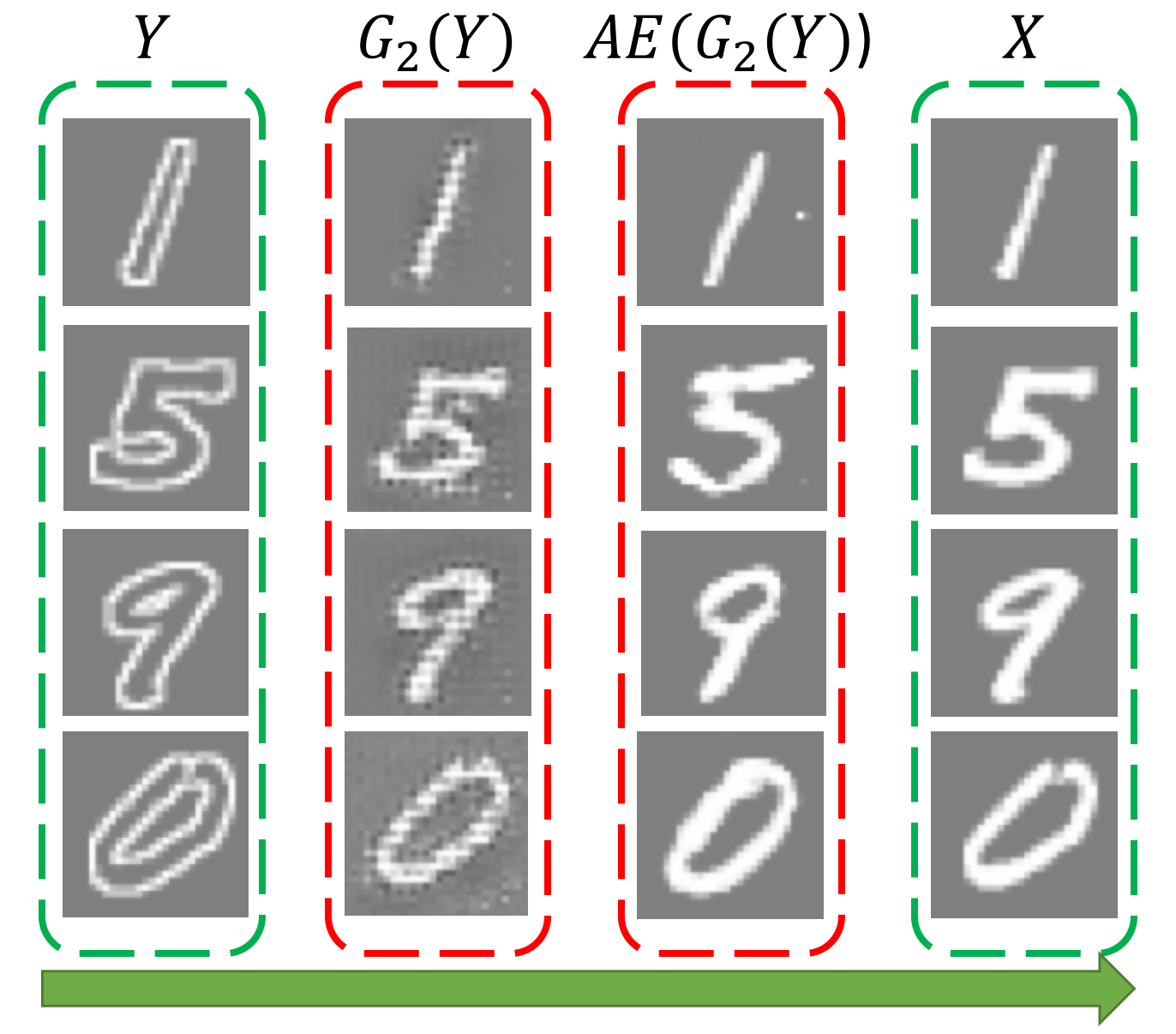}
		\caption{Outputs from ${Y}$ to ${X}$.}
		\label{fig:gull2}
	\end{subfigure}%
	
	\caption{The VIGAN was able to impute bidirectionally regardless of which view was missing.}\label{fig:Bidirection}
\end{figure}
\begin{figure}[htbp]\
	\center
	\includegraphics[scale=.18]{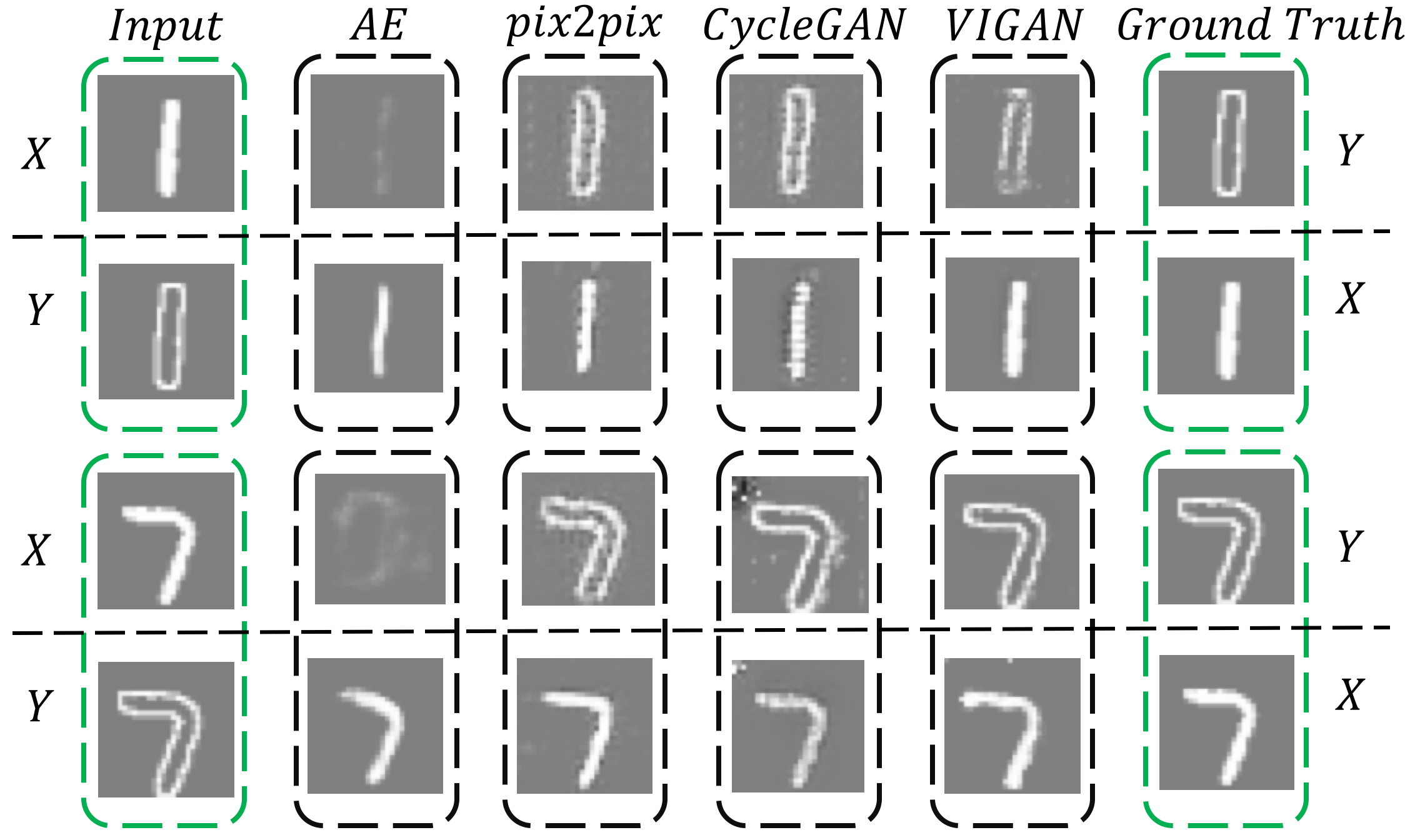}
	\caption{Several examples of $\mathcal{X} \to \mathcal{Y}$ and $\mathcal{Y} \to \mathcal{X}$.} \label{fig:Fige3}
\end{figure}
The images in Figure \ref{fig:Bidirection} illustrate more results.  In both parts of Figure \ref{fig:Bidirection}, the initial view is shown on the left, and the ground truth target is on the right. The two middle columns show the reconstructed images by just the domain mapping, and by the VIGAN.

\noindent \textbf{Paired data vs all data.}
Table \ref{tab1} demonstrates how using both paired and unpaired data could reduce the root mean squared error (RMSE) between the reconstructed image and the original image.  When \textit{all data} were used, the network was trained in the multi-stage fashion described above. The empirical results validated our hypothesis that the proposed VIGAN could further enhance the results from a domain mapping. 
\begin{table}[htbp]
	\caption{The comparison of the root mean squared errors (RMSE) by the four methods in comparison.}
	\begin{center}
		\begin{tabular}{lcccc}
			\hline
			\textbf{ }&\textbf{ }&\multicolumn{3}{c}{\textbf{RMSE}} \\
			\cline{3-5} 
			\textbf{Methods} &\textbf{Data} & \textbf{\textit{V1 $\to$ V2}}& \textbf{\textit{V2 $\to$ V1}}& \textbf{\textit{Average}} \\
			\hline
			
			Multimodal AE& Paired&5.46 &6.12 & 5.79 \\
			pix2pix  & Paired&4.75 &3.49  & 4.12 \\
			CycleGAN &All data$^{\mathrm{*}}$& 4.58 & 3.38 &  3.98 \\
			\textbf{VIGAN}&All data$^{\mathrm{*}}$ &\textbf{4.52} &\textbf{3.16}  & \textbf{3.84}  \\
			\hline
			\multicolumn{4}{l}{$^{\mathrm{*}}$Paired data and Unpaired data.}
		\end{tabular}
		\label{tab1}
	\end{center}
\end{table}

\noindent \textbf{Comparison with other methods.}
For fair comparison, we compared the VIGAN to several potentially most effective imputation methods, including the domain mappings learned respectively by the pix2pix, CycleGAN, and a multi-modal autoencoder methods.  We show both imputation of $X \to Y$ and $Y \to X$ in Figure \ref{fig:Fige3} after running the same number of training epochs, along with the RMSE values in Table \ref{tab1}. As expected, the multi-modal DAE had a difficult time as it could only take paired information, which constituted only a small portion of the data. Although the CycleGAN and pix2pix were comparable with the VIGAN which performed the best, they did not have an effective way to refine the reconstruction from view correspondence. 

\subsection{Healthcare numerical data}
The proposed method can find great utility in many healthcare problems. We applied the VIGAN to a challenging problem encountered when diagnosing and treating substance use disorders (SUDs). To assist the diagnosis of SUDs, the Diagnostic and Statistical Manual version V (DSM-V) \cite{AmericanPsychiatricAssociation2013} describes 11 criteria (symptoms), which can be clustered into four groups: impaired control, social impairment, risk use and pharmacological criteria. In our dataset, subjects who had exposure to a substance (e.g., cocaine) was assessed using the 11 criteria, which led to a diagnosis of cocaine use disorder. For those who had never been exposed to a substance, their symptoms related to the use of this substance were considered unknown, or in other words missing. Due to the comorbidity among different SUDs, many of the clinical manifestations in the different SUDs are similar \cite{hammersley1990criminality,ball2012effectiveness}. Thus, missing diagnostic criteria for one substance use may be inferred from the criteria for the use of another substance. The capability of inferring missing diagnostic criteria is important. For example, subjects have to be excluded from a genome-wide association study because they had no exposure to the investigative substance, even though they used other related substances \cite{Sun2015,Gelernter2014}. By imputing the unreported symptoms for subjects, sample size can be substantially increased which then improves the power of any subsequent analysis. In our experiment, we applied the VIGAN to two datasets: \textit{cocaine-opioid} and \textit{alcohol-cannabis}. The first dataset was used to infer missing cocaine (or opioid) symptoms from known opioid (or cocaine) symptoms. The second dataset was used to infer missing symptoms from the known symptoms between alcohol or cannabis use.  

A total of 12,158 subjects were aggregated from multiple family and case-control based genetic studies of four SUDs, including cocaine use disorder (CUD), opioid use disorder (OUD), alcohol use disorder (AUD) and cannabis use disorder (CUD). Subjects were recruited at five sites: Yale University School of Medicine (N = 5,836, 48.00\%), University of Connecticut Health Center (N = 3,808, 31.32\%), University of Pennsylvania Perelman School of Medicine (N = 1,725, 14.19\%), Medical University of South Carolina (N = 531, 4.37\%), and McLean Hospital (N = 258, 2.12\%). The institutional review board at each site approved the study protocol and informed consent forms. The National Institute on Drug Abuse and the National Institute on Alcohol Abuse and Alcoholism each provided a Certificate of Confidentiality to protect participants. Subjects were paid for their participation. Out of the total 12,158 subjects, there were 8,786 exposed to cocaine or opioid or both, and 12,075 exposed to alcohol or cannabis or both. Sample statistics can be found in Table \ref{tbl:sample}.

\begin{table}[t]
	\caption{Sample size by substance exposure and race.}
	\label{tbl:sample}
	
	\begin{center}
		\vskip -0.1in
		\setlength{\tabcolsep}{2pt}
		\begin{tabular}{lccc}
			\toprule
			&	African American & European American & Other\\
			\midrule
			Cocaine	 & 3,994	& 3,696  & 655 \\
			Opioid	 & 1,496 & 3,034 & 422 \\
			Cocaine or Opioid	 & 4,104 & 3,981 & 695 \\
			Cocaine and Opioid	 & 1,386 & 2,749 & 382 \\
			Alcohol	 & 4,911 & 5,606 & 825 \\
			Cannabis & 4,839 & 5,153 & 794 \\
			Alcohol or Cannabis & 5,333 & 5,842 & 893 \\
			Alcohol and Cannabis & 4,417 & 4,917 & 726 \\			
			\bottomrule
		\end{tabular}
	\end{center}
	\vskip -0.25in
\end{table}

The sample included 2,600 subjects from 1,109 small nuclear families (SNFs) and 9,558 unrelated individuals. The self-reported population distribution of the sample was 48.22\% European-American (EA), 44.27\% African-American (AA), 7.45\% other race. The majority of the sample (58.64\%) was never married; 25.97\% was widowed, separated, or divorced; and 15.35\% was married. Few subjects (0.06\%) had grade school only; 32.99\% had some high school, but no diploma; 25.46\% completed high school only; and 41.27\% received education beyond high school.

Symptoms of all subjects were assessed through administration of the Semi-Structured Assessment for Drug Dependence and Alcoholism (SSADDA), a computer-assisted interview comprised of 26 sections (including sections for individual substance) that yields diagnoses of various SUDs and Axis I psychiatric disorders, as well as antisocial personality disorder \cite{Pierucci-Lagha2005,Pierucci-Lagha2007}. The reliability of the individual diagnosis ranged from $\kappa=0.47-0.60$ for cocaine, $0.56-0.90$ for opioid, $0.53-0.70$ for alcohol, and $0.30-0.55$ for cannabis \cite{Pierucci-Lagha2007}. 

For both datasets, 200 subjects exposed to the two investigative substances were reserved and used as a validation set to determine the optimal number of layers and the number of nodes in each layer. Another set of 300 subjects with both substance exposure was used as a test set to report all our results. All the remaining subjects in the dataset were used to train models. During either validation or testing, we set a view missing and imputed it using the trained VIGAN and data from the other view.

\begin{table}[htbp]
	\caption{Data 1: $View_1=$ Cocaine and $View_2=$ Opioid. Imputation performance was assessed using the Hamming distance that ranged from 0 to 1.}
	\begin{center}
		\begin{tabular}{lcccc}
			\hline
			\textbf{ }&{ }&\multicolumn{3}{c}{\textbf{Accuracy (\%)}} \\
			\cline{3-5} 
			\textbf{Methods} &\textbf{Data} & \textbf{\textit{V1 $\to$ V2}}& \textbf{\textit{V2 $\to$ V1}}& \textbf{\textit{Average}} \\
			\hline
			
			Matrix Completion&Paired&43.85 &48.13 & 45.99  \\
			Multimodal AE&Paired&56.55 &53.72 & 55.14 \\
			pix2pix & Paired&78.27&65.51  &  71.89 \\
			CycleGAN & All data$^{\mathrm{*}}$&78.62 &72.78  &75.70   \\
			\textbf{VIGAN}& All data$^{\mathrm{*}}$ &\textbf{83.82}  &\textbf{76.24}   &\textbf{80.03}  \\
			\hline
			\multicolumn{4}{l}{$^{\mathrm{*}}$Paired data and Unpaired data.}
		\end{tabular}
		\label{tab2}
	\end{center}
\end{table}

\begin{table}[htbp]
	\caption{Data 2: $View_1=$ Alcohol and $View_2=$ Cannabis. Imputation performance was assessed using the Hamming distance that ranged from 0 to 1.}
	\begin{center}
		\begin{tabular}{lcccc}
			\hline
			\textbf{ }&\textbf{ }&\multicolumn{3}{c}{\textbf{Accuracy (\%)}} \\
			\cline{3-5} 
			\textbf{Methods} &\textbf{Data} & \textbf{\textit{V1 $\to$ V2}}& \textbf{\textit{V2 $\to$ V1}}& \textbf{\textit{Average}} \\
			\hline
			
			Matrix Completion& Paired&44.64 &43.02 &43.83  \\
			Multimodal AE& Paired&53.16 &54.22 & 53.69 \\
			pix2pix  & Paired&57.18 &65.05  & 61.12 \\
			CycleGAN &All data$^{\mathrm{*}}$& 56.60 & 67.31 &  61.96 \\
			\textbf{VIGAN}&All data$^{\mathrm{*}}$ &\textbf{58.42} &\textbf{70.58}  & \textbf{64.50}  \\
			\hline
			\multicolumn{4}{l}{$^{\mathrm{*}}$Paired data and Unpaired data.}
		\end{tabular}
		\label{tab3}
	\end{center}
\end{table}

\noindent \textbf{Reconstruction quality.}
Tables \ref{tab2} and \ref{tab3} provide the comparison results among a matrix completion method \cite{wright2009robust}, the multi-modal DAE \cite{ngiam2011multimodal}, pix2pix \cite{isola2016image}  and CycleGAN \cite{zhu2017unpaired}.
For the examples that missed an entire view of data, we observed that the VIGAN was able to recover missing data fairly well. We used the Hamming distance to measure the discrepancy between the observed symptoms (all binary symptoms) and the imputed symptoms.  The Hamming distance calculates the number of changes that need to be made in order to turn string 1 of length $x$ into string 2 of the same length.  Additionally, we observed that the reconstruction accuracy in both directions was consistently higher than that of other methods. Our method also appeared to be more stable regardless of which view to impute.

\noindent \textbf{Paired data vs all data.}
Tables \ref{tab2} and \ref{tab3} show results of the different methods that used paired datasets only such as the multi-modal DAE and pix2pix methods against those  that utilized unpaired data during training.  The results supported our hypothesis that the unpaired data could help improve the view imputation from only the paired data. 

\noindent \textbf{Comparison with CycleGAN.}
Since we used CycleGAN as a basis of the VIGAN, it was important to compare the performance of our method and CycleGAN.  While CycleGAN did a good job for the image-to-image domain transfer problem it struggled in imputing numeric data. We believe that this might be the value that the multi-modal DAE brought additionally to improve accuracy.

\noindent \textbf{Multi-view generalization of the model.}
Although the proposed method was only tested in a bi-modal setting with two views, it can be readily extended to three or more views.  The extension of CycleGAN to a tri-modal setting would be similar to that described by the TripleGAN method \cite{li2017triple}. Extending the VIGAN to more views would also require constructing and pre-training multi-modal autoencoders.   

\noindent \textbf{Scalability.} One of the important advantages of the VIGAN method is its scalability inherited from the use of deep neural networks.  The VIGAN can carry on with very large datasets or a very large amount of parameters due to the scalability and convergence property of the stochastic gradient-based optimization algorithm, i.e. Adam. Imputation of missing values in massive datasets has been impractical with previous matrix completion methods. In our experiments, we observed that matrix completion methods failed to load data into memory, whereas the VIGAN training took only a few hours at most on a Tesla K40 GPU to obtain competitive imputation accuracy.

\section{Conclusion}
We have introduced a new approach to the view imputation problem based on generative adversarial networks which we call the VIGAN. The VIGAN constructs a composite neural network that consists of a cycle-consistent GAN component and a multi-modal autoencoder component, and needs to be trained in a multi-stage fashion. We demonstrate the effectiveness and efficiency of our model empirically on three datasets: an image dataset MNIST, and two healthcare datasets containing numerical vectors.  Experimental results have suggested that the proposed VIGAN method is capable of knowledge integration from the domain
mappings and the view correspondences to effectively recover a missing view for a sample. Future work may include the extension of the existing implementation to more than two views, and its evaluation using additional large datasets from a variety of different domains.  In the future, we also plan to augment the method to be able to identify which view impacts the imputation the most, and consequently, may facilitate the view selection.


\section*{Acknowledgment}
We acknowledge the support of NVIDIA Corporation with the donation of a Tesla K40C GPU. This work was funded by the NIH grants R01DA037349 and K02DA043063, and the NSF grants IIS-1718738 and CCF-1514357.  The authors would like to thank Xia Xiao for helpful discussion, and Xinyu Wang for helping with the experiments.


%


\small
\bibliographystyle{IEEEtran}
\setlength{\bibsep}{0.0pt}
\vskip 0.2in
\bibliography{biblio_chao}

\end{document}